\documentclass[11pt,a4paper]{article}
\usepackage[hyperref]{eacl2021}
\usepackage{times}
\usepackage{latexsym}
\usepackage{graphicx}
\usepackage{amsmath}

\usepackage{microtype}
\usepackage{verbatimbox}
\aclfinalcopy 

\usepackage{etoolbox}
\makeatletter
\patchcmd{\@verbatim}
  {\verbatim@font}
  {\verbatim@font\small}
  {}{}
\makeatother

\title{Bootstrapping Multilingual AMR with Contextual Word Alignments}

\author{Janaki Sheth**\thanks{~~This research was done during an internship at IBM Research AI.} \hspace{1 em}
  Young-Suk Lee* \hspace{1 em}
  Ram\'{o}n Fernandez Astudillo* \hspace{1 em}
  Tahira Naseem*\AND
  Radu Florian* \hspace{1 em}
  Salim Roukos* \hspace{1 em} 
  Todd Ward*\\\\
  **University of Pennsylvania, Philadelphia, PA, USA
  *IBM Research, Yorktown Heights, NY, USA\\
  \texttt{Janaki.Sheth@Pennmedicine.upenn.edu, ramon.astudillo@ibm.com} \\
  \texttt{ysuklee, tnaseem, raduf, roukos, toddward@us.ibm.com}
  }

\date{}

\begin{document}
\maketitle
\begin{abstract}
We develop high performance multilingual Abstract Meaning Representation (AMR) systems by projecting English AMR annotations to other languages with weak supervision. We achieve this goal by bootstrapping transformer-based multilingual word embeddings, in particular those from cross-lingual RoBERTa (XLM-R large). We develop a novel technique for foreign-text-to-English AMR alignment, using the contextual word alignment between English and foreign language tokens. This word alignment is weakly supervised and relies on the contextualized XLM-R word embeddings. We achieve a highly competitive performance that surpasses the best published results for German, Italian, Spanish and Chinese.
\end{abstract}

\section{Introduction}
\label{introduction}

\begin{figure}
    \centering
    \includegraphics[scale=0.5]{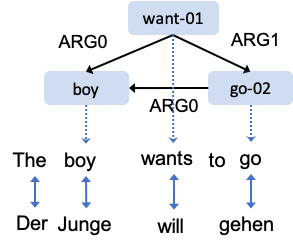}
    \caption{AMR graph for \textit{The boy wants to go} and its German translation \textit{Der Junge will gehen}. Implicit alignments between the English text and AMR concepts are denoted by dotted arrows. Explicit alignments between English and German texts are denoted by solid arrows.}
    \label{fig:amrbasic}
\end{figure}

Abstract Meaning Representation graphs are rooted, labeled, directed, acyclic graphs representing sentence-level semantics \citep{banarescu-etal-2013-abstract}. In the example shown in Figure~\ref{fig:amrbasic}, the sentence \textit{The boy wants to go} is parsed into an AMR graph. The nodes of the AMR graph represent the AMR concepts, which may include normalized surface symbols e.g. \textit{boy}, Propbank frames \cite{kingsbury2002treebank} e.g. \textit{want-01}, \textit{go-02} as well as other AMR-specific constructs. Edges in an AMR graph represent the relations between concepts. In this example \textit{:arg0}, \textit{:arg1} correspond to standard roles of Propbank. 

One distinctive aspect of AMR annotation is the lack of explicit alignments between nodes in the graph and words in the sentences. Since such alignments are essential for training many of present-day AMR parsers, there have been various efforts to link the AMR concepts to their corresponding span of words \citep{flanigan-etal-2014-discriminative, pourdamghani-etal-2014-aligning, lyu-titov-2018-amr, chen-palmer-2017-unsupervised}. A significant emphasis of this paper is on deriving these alignments for multilingual AMR parsers.

Even though by nature AMR is biased towards English, recent work has evaluated the potential of AMR to work as an interlingua. \citet{hajic-etal-2014-comparing} and \citet{ xue-etal-2014-interlingua} categorize and propose refinements for divergences in the annotation between English and Chinese as well as Czech AMRs. \citet{anchieta-pardo-2018-towards} import the corresponding AMR annotation for each sentence from the English annotated corpus and revisit the annotation to adapt it to Portuguese. However, \citet{damonte-cohen-2018-cross} show that it may be possible to use the original AMR annotations devised for English as representation for equivalent sentences in other languages without any modification despite the translation divergence. This defines the problem of multilingual AMR parsing that we seek to address in this paper - given a sentence in a foreign language, recover the AMR graph originally designed for its English translation. We implement multilingual AMR parsers for German, Spanish, Italian and Chinese. 

In this paper we propose that transformer-based multilingual word embeddings can be a useful tool for addressing the problem of multilingual AMR parsing. Besides using contextual word embeddings as input token embeddings, we leverage them for \textit{annotation projection}, where existing AMR annotations for English are projected to a target language by using contextual word alignments. In our experiments, we employ XLM-RoBerta large \citep{Conneau19-xlmr} as the multilingual pre-trained transformer model. We show that our proposed procedure achieves competitive results as some of the classical methods for text-to-AMR alignment. Furthermore, such a procedure is easily scalable to the 100 languages that XLM-R is trained on.

We also combine different techniques for concept alignments and AMR parser training which significantly improve performance over the base models. For concept alignment, we combine the proposed contextual word alignments with previously established alignment techniques utilizing matching rules tailored to AMR as well as machine translation aligners \citep{flanigan-etal-2014-discriminative, pourdamghani-etal-2014-aligning}. For AMR parser training, we pre-train an AMR parser on the treebanks of different languages simultaneously and subsequently finetune on each language. This is analogous to the techniques used for silver data pre-training \citep{konstas-etal-2017-neural, van2017neural} in AMR parsing and multi-lingual pre-training \citep{aharoni-etal-2019-massively} in machine translation. 

Finally, we conduct a detailed error analysis of the multilingual AMR parsing. One of the major errors we have found involves  synonymous concepts, which share the same meaning as the original concepts in English, but differ in spellings. While this error is mainly caused by the fact that the multilingual word embeddings bridge non-English input tokens to English concepts, it also highlights the highly lexical nature of Smatch scoring \citep{cai-knight-2013-smatch} which does not take synonymous concepts into consideration. We also elaborate upon error analysis of the direct comparison between our proposed annotation projection method using contextual word alignment and a previous baseline, using fast align.

The rest of the paper is organized as follows: In Section~\ref{relatedwork}, we discuss related work.  In Section~\ref{method}, we present our main proposal on annotation projection based on contextual word alignments. In Section~\ref{ensemblemethod}, we describe various combination approaches that improve the multilingual parser performances significantly. These include 
combining word-to-concept alignments, using multi-lingual treebanks and combining human-annotated and synthetic treebanks. In Section~\ref{results}, we discuss experimental results. In Sections~\ref{error-analysis} and \ref{word-alignment-error-analysis}, we present detailed error analyses. We conclude the paper in Section~\ref{conclusion}.

\section{Related work}
\label{relatedwork}

\textbf{Multilingual AMR.} There have been significant advances in AMR parsing for languages other than English. Previous studies \citep{hajic-etal-2014-comparing, xue-etal-2014-interlingua, migueles-abraira-etal-2018-annotating, sobrevilla-cabezudo-pardo-2019-towards} investigated AMR annotations for a variety of different languages such as Chinese, Czech, Spanish and Brazilian Portuguese. \citet{vanderwende-etal-2015-amr} automatically parse the logical representation for sentences in Spanish, Italian, German and Japanese, which is then converted to AMR using a small set of rules. 

While much of this work, along with studies such as \citet{li-etal-2016-annotating, anchieta-pardo-2018-towards}, produces AMR graphs whose nodes were labeled with words from the target language, \citet{damonte-cohen-2018-cross} developed AMR parsers for English and used parallel corpora for annotation projection to train Italian, Spanish, German, and Chinese parsers that recover the AMR graph originally designed for the English translation. Their main results showed that the new parsers can overcome certain structural differences between languages. 

Similar to \citet{damonte-cohen-2018-cross}, we also train multilingual AMR parsers by projecting English AMR annotation to target foreign languages (German, Spanish, Italian and Chinese), but we depart from their approach in the specifics of the annotation projection by exploring contextual word alignments directly derived from multilingual contextualized word embeddings. 
While both procedures utilize parallel corpora, the annotation projection of \citet{damonte-cohen-2018-cross} requires additional supervised training of their statistical word aligner. Our proposed contextualized word alignment is however unsupervised in nature. 
Alternatively, a recent study by \citet{blloshmi-etal-2020-xl} showed that one may in fact not need alignment-based parsers for cross-lingual AMR, rather modelling concept identification as a \textit{seq2seq} problem. In this paper, we will compare our results to both \citet{damonte-cohen-2018-cross} and \citet{blloshmi-etal-2020-xl}.

\textbf{Word vector alignment techniques.}  Traditional word alignment methods often use parallel corpora and IBM alignment models \citep{brown1991, brown1993} as well as improved versions \citep{och2003systematic,dyer-etal-2013-simple}. More recently, there have been an advent of techniques that align vector representation of words from varying levels of supervision \citep{Ruder19}. Often word vectors are learned independently for each language and then a mapping from source language vectors to target language vectors with a bilingual dictionary is developed \citep{Mikolov13a, smith17, artetxe-etal-2017-learning}. To reduce the need for bilingual supervision, the iterative method of starting from a minimal seed dictionary and alternating with learning the linear map was employed by a recent body of work \citep{conneau18, schuster-etal-2019-cross, artetxe-etal-2018-robust}. 

The work most similar to ours is \citet{cao20} where the authors obtain contextual embedding alignments from multilingual BERT \citep{devlin2018-bert, pires2019-multilingualbert} and subsequently improve the alignments via finetuning using supervised parallel corpora. Our contextual word alignment between two parallel sentences may be thought of as an adaptation of their contextual word retrieval task. However, we refrain from any finetuning of the contextual embeddings and show that the contextual word alignments from the off-the-shelf XLM-R model achieves results competitive to the word alignments by fast-align (see \citet{damonte-cohen-2018-cross}). This suggests the potential for inexpensive, massive scaling of AMR parsing up to 100 languages on which XLM-R is trained. 

\section{Annotation projection}
\label{method}

We adopt a transition-based parsing approach for AMR parsing following \cite{ballesteros-al-onaizan-2017-amr,naseem-etal-2019-rewarding,astudillo-etal-2020-transition}. These produce an AMR graph $g$ from an input sentence $s$ by predicting instead an action sequence $a$ from $s$ as a sequence to sequence problem. This action sequence applied to a state machine $M$ produces then the desired target graph as $g=M(a,s)$. Transition-based parsers require the action sequence for each graph in the training data. This is determined by a rule-based oracle $a = O(g, s)$ which relies on external word-to-node alignments. In all the subsequent experiments we will use the oracle and action set from \cite{astudillo-etal-2020-transition}.

\subsection{Projection method}

In order to train AMR parsers in a non-English language, we use the annotation projection method to leverage existing English AMR annotation and overcome resource shortage in the target language.  First, the English text is aligned to corresponding AMR concepts using both rule-based JAMR aligner \citep{flanigan-etal-2014-discriminative} and a IBM model type aligner \citep{pourdamghani-etal-2014-aligning}. The latter will henceforth be referred to as the EM aligner. Given the English text-to-AMR concept alignments, we then project these to the target language using word alignment. In the following subsection we describe in the proposed word alignment method, called \textit{contextual word alignment}, which is trained in a weakly supervised manner.

\subsection{Contextual word alignments}
\label{Contextual-word-embeddings}

Given two languages, we align word pairs within parallel sentences if their vector representations derived from the underlying multilingual pre-trained model are similar according to cosine distance. As vector representation we use the average of all 24 layers of the XLM-R large contextual embeddings. We will refer to this average as the word's contextual embedding henceforth for simplicity. 

More precisely, suppose we have two parallel sentences - $\textbf{E} = {e_0, e_1, e_2, ..., e_M}$ in English and $\textbf{F} = {f_0, f_1, f_2, ..., f_N}$ in the target language. We will use $r$ to represent the pre-trained multilingual model such that $r(\textbf{S})_i$ is the contextual embedding for the $i^{th}$ word in sentence $\textbf{S}$. Then a word $e_i \in \textbf{E}$ is contextually word aligned to $f_j$ if and only if the cosine similarity score between their word embeddings is the highest. Thus we define the corresponding contextual alignment function $\chi(f_j|e_i)$ as,
\begin{equation}
    \chi(f_j|e_i) = \mathrm{argmax}_{0 \leq j \leq \bf|F|} cos(r(\textbf{E})_i, r(\textbf{F})_j).
\end{equation}

Similarly, performing the same procedure in the reverse direction we have,
\begin{equation}
    \chi(e_i|f_j) = \mathrm{argmax}_{0 \leq i \leq \bf|E|} cos(r(\textbf{F})_j, r(\textbf{E})_i)
\end{equation}

While these methods can be noisy, by only keeping word pairs in their intersection \textit{i.e.} $\chi(\textbf{E}|\textbf{F})\cap\chi(\textbf{F}|\textbf{E})$, one can derive the intersection cosine alignment approach which gives us a word-aligned dataset with low coverage but high accuracy.

As an example, the following are sentences from our German and English training datasets: \\
\\
\noindent
$\textbf{E:}$ Establishing models in industrial Innovation\\
$\textbf{F:}$ Etablierung von Modellen in der industriellen Innovation\\
Their contextual word alignments are,\\
$\chi(\textbf{F}|\textbf{E})$ = $[(e_0, f_0), (e_1, f_2), (e_2, f_3), (e_3, f_5),\\ (e_4, f_6)]$ \\
$\chi(\textbf{E}|\textbf{F})$ = $[(f_0, e_0), (f_1, e_1), (f_2, e_1), (f_3, e_2), \\
(f_4, e_2), (f_5, e_1), (f_6, e_4)]$\\
$\chi(\textbf{F}|\textbf{E}) \cap \chi(\textbf{E}|\textbf{F})$\\ = $[(e_0, f_0), (e_1, f_2), (e_2, f_3), (e_4, f_6)]$

\begin{figure}
    \centering
    \includegraphics[scale=0.5]{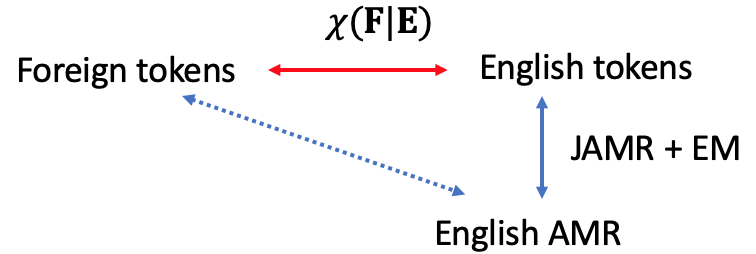}
    \caption{Annotation projection is achieved using JAMR and EM aligners for English text-to-AMR concept alignment and contextual word alignment between tokens of the source (English) and target languages.}
    \label{fig:annotationprojection}
\end{figure}

\vspace{0.2in}
Figure \ref{fig:annotationprojection} pictorially illustrates our complete annotation projection method using the contextual word alignment $\chi(\textbf{F}|\textbf{E})$. English tokens and AMR concepts are aligned using JAMR and EM aligners. The resulting AMR annotation augmented with English word-to-concept alignments is then projected onto the given target language using contextual word embeddings. Henceforth, for brevity we will at times refer to this approach as A.P.

\section{Combination approaches}
\label{ensemblemethod}

We apply three types of combination techniques to the multilingual AMR parsers, trained by projecting English annotations using contextual word alignments derived from the multilingual contextual word embeddings, each of which improves the parser performance significantly.

\begin{figure}
    \centering
    \includegraphics[scale=0.48]{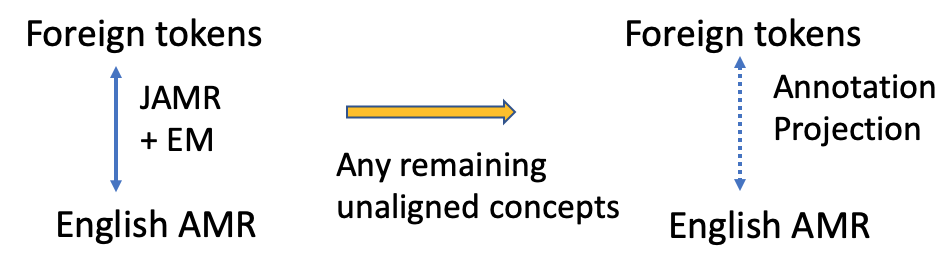}
    \caption{Illustration of the EM, JAMR + A.P. combination alignment: first align target tokens to AMR concepts using JAMR+EM aligners with any remaining concepts then aligned using the annotation projection method proposed in Figure \ref{fig:annotationprojection}.}
    \label{fig:ensemblealignment}
\end{figure}


\subsection{Alignment combination}

One such technique is to combine the contextual word alignment based A.P. with the baseline word-to-concept alignment which aligns the target tokens directly to AMR concepts using JAMR and EM aligners. Since the EM aligner is an unsupervised method, it can be directly applied to the target language tokens and English AMR concepts. However, we note that this baseline alignment approach gives incomplete coverage (87$\%$ concepts aligned to German, 88$\%$ to Italian and 91$\%$ to Spanish tokens). Thus, we supplement this by aligning the remaining concepts using the A.P. of Figure \ref{fig:annotationprojection}. 

For example, suppose we have as before two parallel sentences - $\textbf{E} = {e_0, ..., e_M}$ in English and $\textbf{F} = {f_0, ..., f_N}$ in the target language, as well as AMR concepts ${\textbf{N} = {n_0, ..., n_L}}$. Then one of our proposed foreign text-to-AMR concept combination alignment procedures $EA(f_i|n_j)$ (see Figure \ref{fig:ensemblealignment}) is defined as,
\begin{equation}
\label{eqn_ensemblebaselineap}
    EA(f_i|n_j) = AP(BA(f_i|n_j))
\end{equation}
where $BA(f_i|n_j)$ represents that the $j^{th}$ concept is aligned to the $i^{th}$ token in \textbf{F} using the baseline aligner ${BA}$. If for any concept $n_j \in \textbf{N}$, $BA(f_i|n_j) = \textbf{\rm{None}}$, we use annotation projection to align it where $AP(f_i|n_j)$ is given by,
\begin{equation}
    \chi(f_i|e_k) \wedge BA(e_k|n_j) \Rightarrow AP(f_i|n_j) 
\end{equation}

We also experiment with other such alignments, in particular by using the intersection of cosine alignment ($\chi(F|E) \cap \chi(E|F)$) as the contextual word alignment. In this case,
\begin{equation}
\label{eqn_ensembleintersectionalign}
    EA(f_i|n_j) = \mathrm{max}AP(BA(\mathrm{i}AP(f_i|n_j)))
\end{equation}
wherein,
\begin{equation}
    (\chi(f_i|e_k) \cap \chi(e_k|f_i)) \wedge BA(e_k|n_j) \Rightarrow \mathrm{i}AP(f_i|n_j) 
\end{equation}
As before, $\forall n_j \in \textbf{N}$ where $\mathrm{i}AP(f_i|n_j) = \textbf{\rm{None}}$ we align it using the baseline aligner $BA(f_i|n_j)$. For any further remaining unaligned concepts, we employ $\mathrm{max}AP(f_i|n_j)$ which can be described as:
\begin{equation}
\label{eqn_7}
\begin{split}
    \mathrm{max}(\chi(f_i|e_k), \chi(e_k|f_i)) \wedge BA(e_k|n_j) \\ 
    \Rightarrow \mathrm{max}AP(f_i|n_j) 
\end{split}
\end{equation}

\noindent
That is, we pick the uni-directional contextual word alignment with the higher score and project the AMR annotation accordingly.


\subsection{Multilingual treebank combination}

In addition to training the parser on the treebank of each language - derived from English treebank via annotation projection - we also experiment with combining all the target language treebanks  to create a single multilingual treebank. We notice that pre-training an AMR parser on this multilingual treebank with subsequent finetuning on the treebank of each language, improves performance over the parser trained only on each individual treebank.

\subsection{Human and synthetic treebank combination}

We create a synthetic AMR corpus by parsing 85k unlabeled sentences from the context portion of SQuAD-2.0. The resulting synthetic AMR graphs are filtered as per the procedure in \cite{lee-etal-2020-transition} and combined with the AMR-2.0 training set (LDC2017T10), to produce an expanded \textit{AMR-2.0 + SQuAD} training dataset of 94k sentences. We then project annotations of this expanded English treebank onto each of the target languages, and train the corresponding target language parser. We observe that despite the lower quality of the synthetic AMRs as compared to their human-annotated counterparts, their inclusion in the training set significantly improves parser performance.

\section{Experimental Results}
\label{results}

\subsection{AMR Parser and Data}
\label{Experimental-details}

For our experiments, we use the stack-Transformer model \citep{astudillo-etal-2020-transition}\footnotemark\footnotetext{\url{https://github.com/IBM/transition-amr-parser}} as our AMR parser. The stack-Transformer is a transition based parser with a modified Transformer architecture to encode the parser state. It uses a cross entropy loss function and has hyper-parameters similar to those of machine translation described in \citep{vaswani2017}. We use a beam size of $3$ to decode our models and evaluate them using Smatch scores \citep{cai-knight-2013-smatch}. Model performance values in this manuscript are an average over the best performing models across $3$ random seeds. Lastly, the input to the parser - the vector representation of each word - is obtained by averaging over not only all 24 layers of the pre-trained XLM-R large contextual embeddings but also over constituent wordpieces within each word. 

For all four languages - German, Spanish, Italian and Chinese - we experiment on AMR1.0 (LDC2015E86). For the first three we also experiment on AMR-2.0 (LDC2017T10). Results from the former are compared to \citet{damonte-cohen-2018-cross} and from the latter to \citet{blloshmi-etal-2020-xl}. Details of our training, dev and test sets are given in Table\ref{Data}.\footnote{Word segmentation is applied to the Chinese raw texts for model training and testing.} To train each target language parser, we first translate the input sentences of AMR-2.0 and AMR-1.0 with Watson Language Translator.\footnote{ https://www.ibm.com/watson/services/language-translator/} This creates the supervised parallel corpus which we then use for our unsupervised annotation projection via contextual word alignment. We also align target language tokens directly to AMR concepts using JAMR and EM aligners for baseline system evaluation and for combination alignments. We select the best performing models using the devset. Finally, for our best models, we report results using the machine as well as human translations (LDC2020T07) of the test sets. 

\begin{table*}
\centering
\begin{tabular}{llccccc}
\hline
\textbf{Data set} & \textbf{Experiment} & \textbf{Number of sentences} & \multicolumn{4}{c}{\textbf{Number of tokens}} \\
\hline
& & & DE & ES & IT & ZH \\
\hline
Train set       & AMR2.0 LDC          & 36k & 677k & 694k & 654k & \\
                & AMR2.0 LDC + synAMR & 94k & 2.1m & 2.2m & 2.1m & \\
                & AMR1.0 LDC          & 10k & 222k & 240k & 227k & 195k \\
Development set & All experiments     & 1368 & 30k & 32k & 31k & 26k \\
Test set & All experiments            & 1371 & 31k & 33k & 32k & 27k \\
\hline
\end{tabular}
\caption{\label{Data}
Details of our dataset 
}
\end{table*}

\subsection{Baselines}

Our first baseline is zero-shot learning, where we train on the English dataset but test on a foreign language dev-set (Baseline I). The reason behind this experiment is to test the ability of the XLM-R contextual word embeddings to capture the meaning of the given token irrespective of the underlying language. Note that it is only for this experiment that languages for the train and dev sets differ. 
In another set of experiments we align the target language tokens directly to the AMR concepts only using the JAMR and EM aligners (Baseline II). Lastly, we also test the annotation projection procedure of \citet{damonte-cohen-2018-cross}. Note that while the previous authors use fast align \citep{dyer-etal-2013-simple} for word alignment between the parallel data and only JAMR aligner for the English text-to-AMR alignment, in Baseline III we have utilized fast align in conjunction with both JAMR and EM aligners (for English text-to-AMR alignment) for improved performance.

\begin{table*}
\centering
\begin{tabular}{llllllll}
\hline
\textbf{Model} & \multicolumn{3}{c}{\textbf{AMR2.0}} & \multicolumn{4}{c}{\textbf{AMR1.0}} \\
\hline
& DE & ES & IT & DE & ES & IT & ZH \\
\hline
 Baseline I (zero-shot) & 39.0  & 39.6  & 41.0 & 37.4 & 38.8  & 39.3 & 33.4 \\
 Baseline II  & 61.4  & 66.2  & 68.3 & 57.2 & 60.3  & 60.7 & 55.4\\
 Baseline III  & 63.8  & 68.7  & 68.6 & 56.3 & 60.8 & 61.0 & 54.7 \\
 \hline
 Annotation Projection (A.P) & 61.9 & 67.7 & 66.8 & 55.7 & 60.7 & 60.5 & 46.5  \\
 EM,JAMR+A.P & 63.9 & 68.7 & 69.8 & 57.7 & 62.3 & 62.5 & 55.8 \\
Intersect A.P+EM,JAMR+$\mathrm{max}$(A.P) & 64.2 & 69.1 & 68.7 & \\
EM,JAMR+A.P (Multilingual) & 64.6 & 69.2 & 70.4  & \textbf{58.6} & \textbf{62.7} & \textbf{62.9} & \textbf{58.1}\\
EM,JAMR+A.P (synAMR) & \textbf{67.8} & \textbf{71.3} & \textbf{72.2} \\
\hline
\end{tabular}
\caption{\label{Results2.0}
Dev set Smatch for AMR2.0 and AMR1.0. 
}
\end{table*}


\begin{table*}
\centering
\begin{tabular}{lllllllll}
\hline
\textbf{Model} & \multicolumn{4}{c}{\textbf{Machine translation}} & \multicolumn{4}{c}{\textbf{Human translation}} \\
\hline
& DE & ES & IT & ZH & DE & ES & IT & ZH \\
\hline
\citet{damonte-cohen-2018-cross} & & & & & 39 & 42 & 43 & 35 \\
Baseline I (zero-shot) & 37.1 & 37.99 & 38.5 & 31.8 & 36.3 & 37.6 & 37.4 & 30.2 \\
Baseline II  & 56.1 & 58.94 & 59.7 & 53.3 & 53.6 & 57.8 & 56.8 & 48.3\\
Baseline III  & 55.1 & 59.24 & 59.0 & 53.1 & 52.7 & 57.9 & 57.3 & 48.1 \\
Annotation Projection (A.P) & 54.9 &  58.9 & 59.4 & 44.6 & 52.7 & 57.7 & 57.0 & 41.4 \\
EM,JAMR + A.P & 56.4 & 60.6 & 61.3 & 54.0 & 53.6 & 59.2 & 58.6 & 48.3 \\
EM,JAMR + A.P (Multilingual) & \textbf{57.4} & \textbf{61.4} & \textbf{61.6} & \textbf{55.7} & \textbf{54.5} & \textbf{60.1} & \textbf{59.0} & \textbf{50.3} \\
\hline
\end{tabular}
\caption{\label{Results_Test1.0}
Test set Smatch for AMR1.0. 
}
\end{table*}

\begin{table*}
\centering
\begin{tabular}{llllllll}
\hline
\textbf{Model} & \multicolumn{3}{c}{\textbf{Machine translation}} & \multicolumn{4}{c}{\textbf{Human translation}} \\
\hline
& DE & ES & IT & DE & ES & IT & ZH\\
\hline
\citet{blloshmi-etal-2020-xl} & & & & 53 & 58 & 58.1 & 43.1\\
EM, JAMR + A.P (Multilingual) &  63.8  & 67.7  &  69.0  & 59.9  & 66.0  &  65.7\\
EM, JAMR + A.P (synAMR) & \textbf{66.9}  &  \textbf{69.6} &  \textbf{71.0}  & \textbf{62.7}  &  \textbf{67.9} &  \textbf{67.4}\\
\hline
\end{tabular}
\caption{\label{Results_Test2.0}
Test set Smatch for AMR2.0. 
}
\end{table*}
\subsection{Results}

Table \ref{Results2.0} compares our different proposed approaches to the three baseline methods using the AMR2.0 and AMR1.0 datasets. 
We see that our proposed approach - annotation projection with contextual word alignment, in this case using $\chi(\textbf{F}|\textbf{E})$ - shows fairly competitive results with those of Baseline III for the target languages of German, Italian and Spanish, especially when applied to the smaller corpus of AMR1.0. This is remarkable considering our method requires no additional training and can be easily generalized for zero-shot learning on all different languages that XLM-R was pretrained on. 
We then train several parsers using our suggested combination approaches. The first such method comprises of both the EM, JAMR + A.P aligners (see Eq. \ref{eqn_ensemblebaselineap}). In a different approach, we use the intersection cosine word alignment based annotation projection (\textit{i.e} $\chi(\textbf{F}|\textbf{E}) \cap \chi(\textbf{E}|\textbf{F})$). Since this leaves many AMR concepts unaligned, we follow it by aligning concepts using the baseline JAMR and EM aligners. Any leftover unaligned concepts are then aligned using $\mathrm{max}(\chi(\textbf{E}|\textbf{F}), \chi(\textbf{F}|\textbf{E}))$ (Eq. \ref{eqn_ensembleintersectionalign}). In another set of experiments, we pre-train a parser on a multilingual treebank, where the train set is a combination of the LDC treebank in all target languages. The parser is then finetuned on each individual language. We surmise that such an experiment will give us a truly multilingual parser capable of successfully decoding all the target languages. Its strength is evident in its performance, it outperforms all our baseline approaches - in the case of AMR1.0 dev set by at least 1.4 points. Finally, in the last two experiments on AMR2.0 we train on the language-specific LDC + SQuaD train set. We see that this gives us our best performing parsers, where the training data is aligned using a combination (EM, JAMR + A.P) alignment.

We test a subset of the AMR2.0 and all of the AMR1.0 models on corresponding test sets. The results are shown in Tables~\ref{Results_Test1.0} and~\ref{Results_Test2.0}. For AMR1.0, while all of our models including the baselines outperform previously published results, the best performing model is the parser which was trained on multilingual data and whose training input text was aligned to its AMR concepts using the combination of EM, JAMR and A.P aligners. For AMR2.0, models trained on the LDC + SQuAD dataset out-perform those trained on multilingual data. Both of these outperform the recently published work of \citet{blloshmi-etal-2020-xl}. \footnote{We did not run experiments with LDC + SQuAD dataset on AMR1.0 since our primary reason for running experiments on AMR1.0 was to more directly be able to compare our results to \citep{damonte-cohen-2018-cross}}

We note that the parser performs better on the machine translated test data than on the human translated data. This should be attributed to the training and testing condition mismatch of the human translated test data since all models are trained on machine translated training data. For instance, the out-of-vocabulary (oov) ratio of the human translated test data is consistently higher than that of the machine translated test data. For example, for AMR1.0 the oov ratio of human translated test data vs. machine translated test data is 10.2\% vs. 9\% for German, 7.3\% vs. 6.8\% for Spanish, 8.1\% vs. 7.6\% for Italian and 7.6\% vs. 5.5\% for Chinese.



\begin{figure}
    \centering
    \includegraphics[scale=0.3]{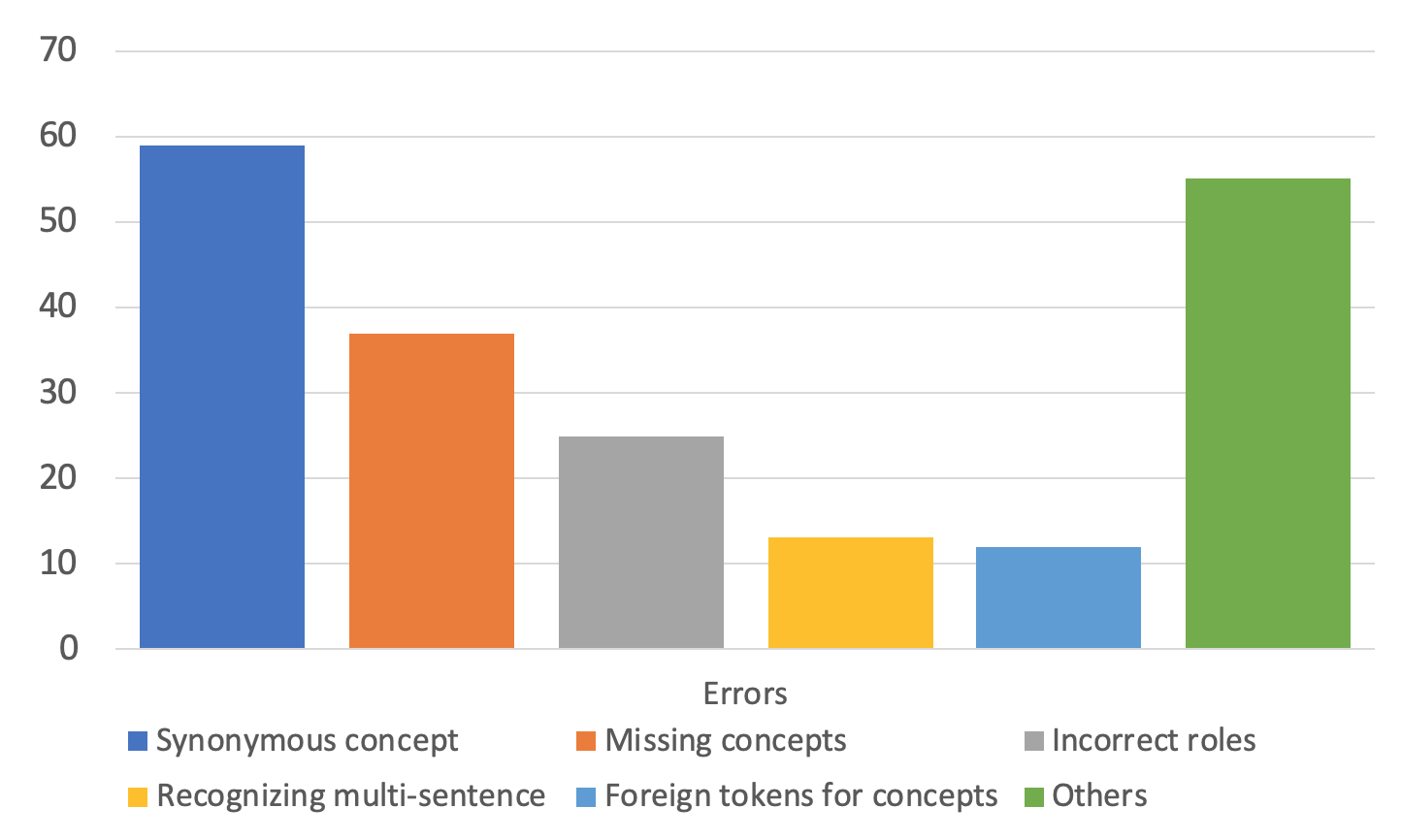}
    \caption{Histogram of different kinds of errors}
    \label{fig:errorhistogram}
\end{figure}

\section{Error analysis}
\label{error-analysis}

We carried out an error analysis of 56 German sentences parsed by the best performing model trained on the combination of AMR2.0 and SQuAD training data. Statistics of the various errors are depicted in Figure \ref{fig:errorhistogram}. Top 5 most frequent errors include (i) introduction of synonymous concepts, (ii) missing concepts, (iii) incorrect roles, (iv) target tokens in AMR concepts, (v) incorrect parsing of multi-sentence as an instance of conjunction.

\subsection{Synonymous concepts}

The most common error we encounter is synonymous AMR concepts, as shown in Figure \ref{fig:synonymerror}. Comparing the expected graph (top) to the parsed version (bottom), we note that concept \textit{previous} is synonymized to \textit{past}. While this error is mainly caused by the fact that the multilingual word embeddings bridge non-English input tokens to English concepts, it also highlights the highly lexical nature of Smatch scoring \citep{cai-knight-2013-smatch} which does not take synonymous concepts into consideration.
Given that AMR is supposed to represent the core meaning of a sentence regardless of its syntactic and morphological variations, Smatch scoring should be able to capture lexical variations such as synonymous concepts.

\begin{figure}
\begin{verbnobox}[\fontsize{8pt}{8pt}\selectfont]  
In this environment, what's wrong if they 
criticize the previous stupefying propaganda 
a bit?
\end{verbnobox}
\vspace{0.02in}
\begin{verbnobox}[\fontsize{8pt}{8pt}\selectfont]  
(w / wrong-02
      :ARG1 (a2 / amr-unknown)
      :ARG2 (c / criticize-01
            :ARG0 (t / they)
            :ARG1 (p / propaganda
                  :time (p2 / previous)
                  :ARG1-of (s / stupefy-01))
            :degree (b / bit))
      :location (e / environment
            :mod (t2 / this)))
\end{verbnobox}
\vspace{0.02in}
\begin{verbnobox}[\fontsize{8pt}{8pt}\selectfont]
Was ist in dieser Umgebung falsch, wenn sie 
die bisherige stupeftende Propaganda ein 
bisschen kritisieren?
\end{verbnobox}
\vspace{0.02in}
\begin{verbnobox}[\fontsize{8pt}{8pt}\selectfont]
(w / wrong-02
      :ARG1 (c / criticize-01
            :ARG0 (t2 / they)
            :ARG1 (p2 / propaganda
                  :time (p / past))
                  :degree (b / bit))
      :ARG2 (a / amr-unknown)
      :location (e / environment
            :mod (t / this)))
\end{verbnobox}    
\caption{The gold AMR (top) and the parsed AMR (bottom) for a German sentence exemplifying errors: synonymous concept (\textit{previous} vs. \textit{past}), missing concept (concept
\textit{stupefy-01} is missing in the parsed AMR), incorrect roles (the two arguments, \textit{:ARG1} and \textit{:ARG2}, of \textit{wrong-02} are swapped in the parsed AMR).}
\label{fig:synonymerror}
\end{figure}

\begin{figure}
\begin{verbnobox}[\fontsize{8pt}{8pt}\selectfont]
In critical moments, we are all descendants of 
Yan emperor and Huang emperor.
\end{verbnobox}
\vspace{0.02in}
\begin{verbnobox}[\fontsize{8pt}{8pt}\selectfont]
(d / descend-01
      :ARG0 (w / we
            :mod (a / all))
      :source (a2 / and
            :op1 (p / person 
                 :name (n / name 
                       :op1 "Yan")
                 :ARG0-of (h / have-org-role-91
                          :ARG2 (e / emperor)))
            :op2 (p2 / person 
                 :name (n2 / name :op1 "Huang")
                 :ARG0-of h))
      :time (m / moment
            :ARG1-of (c / critical-02)))
\end{verbnobox}
\vspace{0.02in}
\begin{verbnobox}[\fontsize{8pt}{8pt}\selectfont]
In kritischen Momenten sind wir alle Nachfahren 
des Yan Kaisers und Huang Kaisers.
\end{verbnobox}
\vspace{0.02in}
\begin{verbnobox}[\fontsize{8pt}{8pt}\selectfont]
(d / descend-01
      :ARG0 (w / we
            :mod (a / all))
      :ARG1 (a2 / and
            :op1 (p / person
                  :name (n / name
                        :op1 "Yan"
                        :op2 "Kaisers"))
            :op2 (p2 / person
                  :name (n2 / name
                        :op1 "Huang"
                        :op2 "Kaisers")))
      :time (m / moment
            :ARG1-of (c / critical-02)))
\end{verbnobox}
\caption{The gold AMR (top) and the parsed AMR (bottom) for a German sentence illustrating incorrect roles (\textit{:source} is replaced by \textit{:ARG1} in the parsed AMR) and incorrect
identification of the target token \textit{Kaisers} as a named entity.}
\label{fig:foreigntokenerror}
\end{figure}

\subsection{Missing concepts and incorrect roles}

Some concepts are missing in the parsed AMR, such as \textit{stupefy-01} in Figure~\ref{fig:synonymerror}. The parser also incorrectly identifies relations between concepts. In Figure \ref{fig:synonymerror}, arguments \textit{ARG1} and \textit{ARG2} for concept \textit{wrong-02} are swapped. In Figure \ref{fig:foreigntokenerror}, the relation \textit{:source} is replaced by frame argument \textit{ARG1}.

\subsection{Incorrect parsing of Multi-sentence}

Another frequent error includes incorrect parsing of multi-sentence as an instance of conjunction, especially when sentences are demarcated by commas. Note that the multi-sentence errors are not specific to multilingual parsing and occur frequently when parsing English input sentences as well. This multi-sentence error is mostly caused by the ambiguity of commas, which can subsume various semantics depending on the contexts across languages.

\subsection{Misrecognition of foreign token as a named entity}

Some target tokens may legitimately be realized in the gold AMR, especially when the target tokens are named entities, e.g. \textit{Frankfurt, Anna, Noah, etc}. This often leads to errors in the parsed AMR when a target token is incorrectly recognized as a named entity. 
In Figure~\ref{fig:foreigntokenerror}, German token \textit{Kaisers} is incorrectly parsed as part of named entities \textit{Yan Kaisers} and \textit{Huang Kaisers}. The failure to capture the correct concept \textit{emperor} for the German token \textit{Kaisers} leads to a subsequent error of not reifying the role to \textit{have-org-role-91}\footnote{Refer to https://www.isi.edu/~ulf/amr/lib/roles.html and https://www.isi.edu/~ulf/amr/lib/amr-dict.html/have-org-role-91 for details.}, evident in the comparison of the parsed AMR with the gold AMRs.

\subsection{Others}

Other errors include lack of stemming in the target language, such as \textit{Kaisers} in 
Figure~\ref{fig:foreigntokenerror}. Stemming errors are mostly caused by the fact that we have not incorporated target language stemmers whereas we have incorporated spacy\footnote{https://spacy.io/} for English. Some errors are caused by machine translation. English fragmentary input \textit{taking a look} is translated to \textit{Sehen Sie sich}, which is then incorrectly parsed as \textit{imperative} sentence. Nominal target language tokens often fail to invoke predicates. Given the input in English  ``cultural tyranny in the cloak of nationalism'', \textit{tyranny} invokes the predicate \textit{tyrannize-01}. Its German counterpart \textit{Tyrannei}, however, fails to invoke the predicate in ``kulturellen Tyrannei im Mantel des Nationalismus''. 

\section{Word alignment error analysis}
\label{word-alignment-error-analysis}
We compared the annotation projection for AMR1.0 between fast align and the contextual alignment. As noted in Table \ref{Results_Test1.0} they perform comparably for German, Italian and Spanish. However, on detailed analysis we notice that annotation projection using contextualized alignments has a greater coverage in terms of foreign text-to-AMR alignments compared to fast align (eg. for German, contextual alignment A.P. gives 99.95\% coverage in comparison to 97.47\%.). This is likely due to the fact that fast align is based on an IBM alignment model, which relies on expected counts of alignment pairs and uses additional alignment constraints. Contextualized alignment relies on the unrestricted pairing by cosine distance of the XLM-R contextual word embeddings of the input tokens. Given an English token, the contextualized alignment necessarily aligns it to a foreign language word. Furthermore, since embeddings are contextual and pre-trained with large amounts of data, they are robust to non frequent alignment pairs. 

The difference between contextualized alignment and fast align for their coverage is most noticeable for compounds. A German counterpart of English non – tariff is \textit{nichttarifäre}. While contextualized alignment aligns \textit{nichttarifäre} to non, which is subsequently aligned to the concept ``–'' for polarity, fast align leaves \textit{nichttarifäre} unaligned. Such difference is evidenced in the parser performance on negations realized in diverse morphologies. Comparing the AMR1.0 parser performance on negations between fast align (Baseline III in Table 3) and the contextualized alignment (A.P in Table 3), we find that contextualized alignment consistently outperforms fast align across the three European target languages, as shown in Table \ref{compounds-align-diff}.

\begin{table}
\centering
\begin{tabular}{lll}
\hline
 & \textbf{Contextual} & \textbf{Fast Align} \\
 & \textbf{Alignment} & \\
\hline
German & 23.47 & 20.52 \\
Italian & 29.40 & 29.30 \\
Spanish & 28.81 & 26.69 \\
\hline
\end{tabular}
\caption{\label{compounds-align-diff}
AMR1.0 parser performance on negations in terms of Smatch. Fast align is compared with the proposed contextual alignment for different languages.
}
\end{table}

\section{Conclusion and future directions}
\label{conclusion}

In this paper we propose to use transformer-based multilingual word embeddings for \textit{annotation projection} of AMR annotations. We show that our proposed procedure achieves competitive results as some of the classical methods for text-to-AMR alignment. We apply combination techniques to concept alignments and AMR parser training, which significantly improve performance over the base models. We also provide a detailed error analysis of the multilingual AMR parsing. 

Given pre-trained transformer-based multilingual word embeddings, contextual word alignment proves to be a useful avenue for overcoming differences amongst languages and addressing the multilingual AMR problem with weak supervision. Moreover, our annotation projection procedure not only achieves a highly competitive performance for German, Spanish, Italian and Chinese but also permits zero-shot learning to other languages included in the training set of the underlying XLM-R multilingual transformer. 

Future work may include diversifying input texts using AMR2text ~\cite{mager-etal-2020-gpt-too} generation which can address the difference in results between machine translated and human translated test data. The potential of the AMR parser to overcome translation divergence also points to its utility in an end-to-end multilingual translation system, bypassing the need for supervised parallel corpora for machine translation system training. 


\section*{Acknowledgement}
We thank the anonymous reviewers for helpful suggestions. We also thank Revanth Reddy and Jason Furmanek for their varied inputs. 

\bibliography{revisedpaper}
\bibliographystyle{acl_natbib}

\end{document}